\documentclass[10pt,letterpaper]{article}

\usepackage{cogsci}
\usepackage{pslatex}
\usepackage{apacite}

\usepackage{amsmath}
\usepackage{amsfonts}
\usepackage{amssymb}
\usepackage{graphicx}
\usepackage[round]{natbib}
\usepackage{color}

\newcommand{\dbc}[2]{\left\langle{#1},{#2}\right\rangle}
\newcommand{\sdbc}[1]{\left\langle{#1}\right\rangle}

\newcommand{\btxt}[1]{\textcolor{black}{#1}}
\newcommand{\rtxt}[1]{\textcolor{black}{#1}}
\newcommand{\tb}[1]{\textcolor{black}{#1}} 
\newcommand{\tbbb}[1]{\textcolor{black}{#1}}

\newcommand{\bb}[1]{\textbf{#1}}

\newcommand{\mc}[1]{\mathcal{#1}}

\newcommand{\gb}{{\color{green}\bullet}}
\newcommand{\rb}{{\color{red}\bullet}}
\newcommand{\wb}{{\circ}}
\newcommand{\indd}{\perp\hspace*{-6pt}\perp}
\newcommand{\inddd}{\perp\hspace*{-4pt}\perp}
\newcommand{\var}{\varnothing}

\title{A Rational Distributed Process-level Account of Independence Judgment}

\author{{\large \bf Ardavan~S.~Nobandegani$^{1,2}$\quad \quad Ioannis~N.~Psaromiligkos$^{1}$}\\ \vspace*{2pt}
\{ardavan.salehinobandegani@mail.mcgill.ca, ioannis.psaromiligkos@mcgill.ca\}\\ \vspace*{-2pt}
\small{$^{1}$Department of Electrical \& Computer Engineering, McGill University}\\
\small{$^{2}$Department of Psychology, McGill University}}

\begin{document}
\maketitle

\begin{abstract}
It is inconceivable how chaotic the world would look to humans, faced with innumerable decisions a day to be made under uncertainty, had they been lacking the capacity to distinguish the relevant from the irrelevant---a capacity which computationally amounts to handling probabilistic independence relations. The highly parallel and distributed computational machinery of the brain suggests that a satisfying process-level account of human independence judgment should also mimic these features. In this work, we present the first rational, \emph{distributed}, message-passing, process-level account of independence judgment, called $\mc D^\ast$. Interestingly, $\mc D^\ast$ shows a curious, but normatively justified tendency for quick detection of dependencies, whenever they hold. Furthermore, $\mc D^\ast$ outperforms all the previously proposed algorithms in the AI literature in terms of worst-case running time, and a salient aspect of it is supported by recent work in neuroscience investigating possible implementations of Bayes nets at the neural level. $\mc D^\ast$ nicely exemplifies how the pursuit of cognitive plausibility can lead to the discovery of state-of-the-art algorithms with appealing properties, and its simplicity makes $\mc D^\ast$ potentially a good candidate for
pedagogical purposes.

\end{abstract}

\section{Introduction}
Is there any connection between the quality of your last night sleep and the color of the shirt your colleague happened to be wearing at work today? How about Mars' current weather and your mood today? We humans judge innumerable such possible connections a day rather effortlessly, appearing to be quite good at teasing apart pertinent from impertinent factors when making decisions. But how does the mind do that? The famous frame problem {(Icard \& Goodman, 2015; Nobandegani \& Psaromiligkos, 2017)}, a puzzle in philosophy of mind and epistemology, further highlights this intriguing ability of the mind in distinguish the relevant from the irrelevant, and asks a closely related question: ``How do we account for our apparent ability to make decisions on the basis only of what is relevant to an ongoing situation without having explicitly to consider all that is not relevant?" (Stanford Encyclopedia of Philosophy). Computationally, the mind's ability of distinguishing the relevant from irrelevant can be characterized in terms of handling probabilistic (in)dependence relations, with `dependency' implying the existence of connection or relevance between factors and `independence' the contrary \citep{pearl1986fusion,pearl1988probabilistic,pearl2000causality}. For example, assuming that the random variable $\bb x$ encodes the quality of your sleep, and $\bb y$ the color of the shirt your colleague happened to wear the next day, the nonexistence of any connection between $\bb x$ and $\bb y$ (which seems to be a rational judgment) can be formally characterized using the notion of probabilistic independence: $\bb x \indd \bb y$ (read $\bb x$ is independent of $\bb y$).

In this work, we are concerned with developing a plausible, process-level account of human independence judgment. Adopting causal Bayes nets (CBNs) (Pearl, 1988; Gopnik et al., 2004, \emph{inter alia}) as a normative model to represent how the reasoner's internal causal model of the world is structured (i.e., reasoner's mental model), the aforesaid task computationally amounts to checking for independencies in the distribution encoded by a CBN. Interestingly, \citet{pearl1986fusion} put forth a graph-theoretic notion called $d$-separation, allowing for reading off probabilistic independence relations from the mere structure of a CBN \citep{pearl1986fusion}.\footnote{More accurately, Pearl's \citeyearpar{pearl1986fusion} $d$-separation is equally valid for Bayes nets wherein the edges do not enjoy causal interpretations.} Ever since its inception, $d$-separation has proved fundamental in a variety of domains in artificial intelligence, e.g., probabilistic reasoning \citep{pearl1988probabilistic}, causal reasoning \citep{pearl2000causality}, decision making \citep{shachter1998bayes,koller2009probabilistic}, and has played important roles in a broad range of areas, e.g., handling missing data \citep{mohan2014testability}, extrapolation across populations \citep{pearl2014external}, and deep learning \citep{goodfellow2016deep}. In that light, algorithms for implementing $d$-separation could potentially serve as a rational, process-level model of human independence judgment. But what should such a model look like? The highly parallel and distributed computational machinery of the brain suggests that a satisfying process-level account of human independence judgment should also mimic these features. Sadly enough, all past algorithms for the implementation of $d$-separation have been \emph{sequential} (aka \emph{serial}), i.e., without any parallelism in computation, and, arguably worse, \emph{centralized}, i.e., their executions are fully coordinated by a supervisory unit, analogous to homunculus \citep{geiger1989d,lauritzen1990independence,shachter1998bayes,koller2009probabilistic,butz2016relevant}, features that strongly call into question their psychological plausibility.

The notion of (conditional) probabilistic independence is a quintessential feature of CBNs, and, interestingly, the realization that probabilistic independence plays a crucial role in human cognition was a key element in the development of the CBN formalism \citep{pearl1986fusion}. In Pearl's \citeyearpar{pearl1986fusion} words: {``Whereas a person may show reluctance to giving a numerical estimate for a conditional probability $P(\bb x_i|\bb x_j)$, that person can usually state with ease whether $\bb x_i$ and $\bb x_j$ are dependent or independent, namely, whether or not knowing the truth of $\bb x_i$ will alter the belief in $\bb x_j$."} He then continues: {``Likewise, people tend to judge the three-place relationships of conditional dependency (i.e., $\bb x_i$ influences $\bb x_j$ given $\bb x_k$) with clarity, conviction, and consistency. This suggests that the notions of dependence and conditional dependence are more basic to human reasoning than are the numerical values attached to probability judgments."} Some psychological literature, however, does not fully embrace the statement ``with clarity, conviction, and consistency" as Pearl put it. For example, the experimental work by Rehder \citeyearpar{rehder2014independence} suggests that adults exhibit deviations from the Markov condition (i.e., CBN's independencies entailed by $d$-separation). In contrast, drawing on the experimental studies of Park and Sloman \citeyearpar{park2013mechanistic}, Sloman and Lagnado \citeyearpar{sloman2015causality} conclude that people indeed uphold the Markov condition and the reason behind the observed deviations is that, under experimental conditions, people may not solely adhere to the information provided by the experimenter and may bring their own background knowledge into the experiment (see also \citealp{Rehder2015}). Specifically, Park and Sloman \citeyearpar{park2013mechanistic} found strong support for their contradiction hypothesis followed by the mediating mechanism hypothesis, and finally concluded that people do conform to Markov condition once the causal structure people are using is correctly specified (i.e., people's mental causal models).

In this work, we present the first rational, \emph{distributed}, process-level account of independence judgment, called $\mc D^\ast$. More formally, $\mc D^\ast$ is the first asynchronous, message-passing, distributed algorithm for implementing $d$-separation, with substantial parallelism in computation, and without any need for a supervisory unit to coordinate its execution (i.e., no synchrony is assumed in $\mc D^\ast$'s execution)---fully in the spirit of the celebrated parallel distributed processing (PDP) research program in brain and cognitive sciences  \citep{McClelland1989}. Similar to the well-known belief propagation inference algorithm \citep{pearl1986fusion,pearl1988probabilistic}, which has played important roles in the theoretical neuroscience literature (see e.g.,  \citealp{gershman2016complex,george2009towards,litvak2009cortical,rao2004bayesian,lochmann2011neural}), $\mc D^\ast$ is a message-passing algorithm, wherein computation is carried out by propagating messages between computational units. Interestingly, $\mc D^\ast$ shows a curious, normatively justified tendency for quick detection of probabilistic dependencies, whenever they hold. Furthermore, $\mc D^\ast$ outperforms all the previously proposed algorithms in the AI literature in terms of worst-case running time, and a salient aspect of it is supported by recent work in neuroscience investigating possible implementations of Bayes nets at the neural level (e.g., \citealp{gershman2016complex,lochmann2011neural}). 

We provide a comprehensive analysis of the computational properties of $\mc D^\ast$, along with several refined time-complexity bounds. In the Discussion section, we provide a detailed comparison between $\mc D^\ast$ and previously proposed algorithms, and elaborate on the implications of the work presented here for neuroscience and psychology.

\section{Preliminaries and Notations}
\label{sec_notation}
Let us introduce the notation adopted in this work. Lower bold-faced letters (e.g., $\bb x$) denote random variables and upper bold-faced letters (e.g., $\bb X$) represent sets of random variables. A generic $d$-separation relation is denoted by $(\bb A \indd \bb B|\bb C)_G$ with $\bb A, \bb B$, and $\bb C$ representing three mutually disjoint sets of variables belonging to the directed acyclic graph (DAG) $G$, where $G$ represents the topology of the underlying CBN. Read $(\bb A \indd \bb B|\bb C)_G$ as follows: $\bb C$ $d$-separates $\bb A$ from $\bb B$ in DAG $G$. Similarly, $(\bb A \not\indd \bb B|\bb C)_G$ denotes that $\bb C$ does not $d$-separate $\bb A$ from $\bb B$ in DAG $G$. For ease of notation, we use $(\bb A \indd \bb B|\bb C)_G$ to denote both a $d$-separation relation (i.e., $\bb C$ $d$-separates $\bb A$ from $\bb B$ in DAG $G$) and to denote a $d$-separation query (i.e., does $\bb C$ $d$-separate $\bb A$ from $\bb B$ in DAG $G$?); the distinction should be clear from the context. Let also $G_{An(\bb K)}$ denote the ancestral graph for the variables in set $\bb K$ belonging to the underlying DAG $G$ \citep{lauritzen1990independence}, i.e., the set of nodes for $G_{An(\bb K)}$ comprises the nodes in $\bb K$ and all the ancestors of the nodes in $\bb K$ (hence, $G_{An(\bb K)}$ is an induced subgraph of the underlying DAG $G$). 

Informally speaking, throughout that paper, $(\bb A \indd \bb B|\bb C)_G$ should be interpreted as follows: ``$\bb A$ and $\bb B$ are probabilistically independent of each other, given $\bb C$," and, in the query format, as follows: ``Are $\bb A$ and $\bb B$  probabilistically independent of each other, given $\bb C$?" Likewise, $(\bb A \not\indd \bb B|\bb C)_G$ should be interpreted as follows: $\bb A$ and $\bb B$ are dependent, given $\bb C$.\footnote{Formally, the said interpretations are not fully granted; however, for all purposes of this work, they can be taken to be accurate enough characterizations (see \tb{Pearl, 2000}, for a complete elaboration on the precise relation between $d$-separation and conditional independence.)}

Next, a notion called refutation-module is introduced; this will be used later in our formal analysis of $\mc D^\ast$.

\begin{figure}[h!]
\centering
\includegraphics[width=0.47\textwidth]{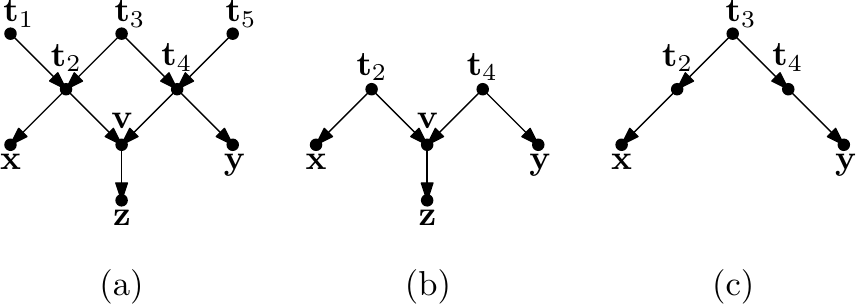}
\caption{\small{Examples for refutation modules. {(a)} The underlying DAG $G$ is depicted, for which $(\bb x \not\indd \bb y|\bb z)_G$. {(b,c)} Two refutation-modules for the \rtxt{query} $(\bb x \indd \bb y|\bb z)_G$ are depicted.}}
\label{fig_ref_module}
\end{figure}

\textbf{Def.~1. (Refutation-Module)} Let $\bb X,\bb Y,\bb Z$ be three mutually disjoint sets belonging to a DAG $G$. Let also $(\bb X \not\indd \bb Y|\bb Z)_G$. A connected subgraph of $G$, $\mc M_{(\bb X \not\inddd \bb Y|\bb Z)_G}$, serves as a refutation-module for the \rtxt{query} $(\bb X \indd \bb Y|\bb Z)_G$, iff $\mc M_{(\bb X \not\inddd \bb Y|\bb Z)_G}$ satisfies the following two conditions: (1) $\mc M_{(\bb X \not\inddd \bb Y|\bb Z)_G}$ contains an active path $P$ \tb{\citep{pearl1986fusion}} between a node $\bb x \in \bb X$ and a node $\bb y \in \bb Y$, and (2) for every head-to-head node $\bb v$ on $P$, $\mc M_{(\bb X \not\indd \bb Y|\bb Z)_G}$ contains a directed path between $\bb v$ and a node $\bb c \in \bb C$. \tb{See Fig.~\ref{fig_ref_module} for some examples.} 

\textbf{Def.~2. (Minimal Refutation-Module)} Let $\bb A,\bb B,\bb C$ be three disjoint sets of nodes belonging to a DAG $G$. Also, let $(\bb X \not\indd \bb Y|\bb Z)_G$. Let $\mc M_{(\bb X \not\indd \bb Y|\bb Z)_G}^{\ast}$ denote the refutation-module for the $d$-separation query $(\bb X \indd \bb Y|\bb Z)_G$ which possesses the smallest number of edges. We refer to $\mc M_{(\bb X \not\indd \bb Y|\bb Z)_G}^{\ast}$ as the \emph{minimal} refutation-module in $G$ for the query $(\bb X \indd \bb Y|\bb Z)_G$. 

It is easy to prove by construction that the minimal refutation-module $\mc M_{(\bb X \not\indd \bb Y|\bb Z)_G}^{\ast}$ need not be unique.

\section{The Three-Color Algorithm $\mc D^\ast$} 
\label{sec_alg_main}
\label{sec_alg_d_star}
In this section, we show how the proposed algorithm $\mc D^\ast$ allows us to decide if a generic $d$-separation query of the form $(\bb A\indd \bb B|\bb C)_G$ holds in a DAG $G$; $\mc D^\ast$ is an asynchronous, distributed, message-passing algorithm. More specifically, in $\mc D^\ast$, nodes of the underlying DAG $G$---symbolizing computational units---autonomously engage in communicating messages to their {immediate} neighbors via the edges of the DAG $G$---symbolizing communication channels. We assume that communication channels are reliable, bidirectional, and first-in first-out (FIFO) \citep{lynch1996distributed}. 

The proposed algorithm $\mc D^\ast$ is outlined next. Throughout an execution of $\mc D^\ast$, variables in $\bb C$ ignore all messages received from any of their children, and do not send any message to any of their children. The variables in the sets $\bb A$, $\bb B,$ and $\bb C$ initially activate in the states represented by colors green ($\color{green}\bullet$), red ($\color{red}\bullet$), and white ($\circ$), respectively. Following the prescriptions of the original Belief Propagation algorithm \citep[Sections 1.3 and 2.2.3]{pearl1986fusion}, we assume that the variables in the sets $\bb A,\bb B,\bb C$ acquire their initial states in a \emph{self-activated} manner.\footnote{Alternatively, we provide an asynchronous, distributed, $O(l)$-time message-passing algorithm in \tb{Sec.~C-VII} of Appendix C, which permits a predesignated source node to disseminate information regarding the initial states of the nodes through the graph $G$, where $l$ denotes the length of the longest undirected path in $G$.} {\tb{Assuming that a CBN's node can be represented at the neural level by a single \citep{deneve2008bb1,deneve2008bb2} or a population of neurons \citep{ma2006bayesian}, self-activation reflects the content-addressability of the corresponding memory traces.}} $\mc D^\ast$ begins with nodes in $\bb A, \bb B$, and $\bb C$ sending their colors as messages to their parents. Node $\bb x$, upon receiving a message, follows two simple steps in the following order:

\begin{itemize}
\item[(i)] If $\bb x$'s current color differs from that of the received message, $\bb x$ replies by sending back its own color as a message to the transmitter node. If $\bb x$ is in the state of having no color (denoted by $\var$) prior to the receipt of the message, it does not send back any message to the transmitter node.
\item[(ii)] $\bb x$ updates its color in accord with the following primitive rules, altogether composing the Color Update Grammar (CUG): 
\begin{eqnarray*}
&&(\var,\gb)\rightarrow \gb, (\var,\rb)\rightarrow \rb, (\var,\wb)\rightarrow \wb,\\
&&(\gb,\gb)\rightarrow \gb, (\rb,\rb)\rightarrow \rb, (\wb,\wb)\rightarrow \wb,\\
&&(\wb,\gb)\rightarrow \gb, (\wb,\rb)\rightarrow \rb,\\
&&(\gb,\wb)\rightarrow \gb, (\rb,\wb)\rightarrow \rb,\\
&&(\gb,\rb)\rightarrow \text{clash}, (\rb,\gb)\rightarrow \text{clash},
\end{eqnarray*}
where the syntax is: ($\bb x$'s current color, received message) $\rightarrow$ $\bb x$'s new color. If $\bb x$'s new color turns out to be different from its old color, \tb{with the exception of the transmitter node}, $\bb x$ sends its new color as a message to all its parents, and only those children of $\bb x$ with which $\bb x$ has communicated before.
\end{itemize}

The rules given in the first row of the CUG correspond to white-, green-, and red-colored nodes sending their colors to their yet-uncolored parents. Rules in the second row ensure that the colors of white-, green-, and red-colored nodes persist upon interacting with nodes of the same color. Rules stated in the third row bear on the key understanding that the white color functions as a mere place-holder getting ``replaced" by interacting with green-, or red-colored nodes. Rules in the fourth row guarantee the persistence of colors green and red  upon interacting with white. Finally, rules given in the last row correspond to the clash event the implication of which is discussed in Remark~1 below.

\textbf{Remark~1.} A clash between colors green ($\color{green}\bullet$) and red ($\color{red}\bullet$) at a node, any time throughout an execution of $\mathcal{D}^\ast$, signals the falsity of the input $d$-separation query, upon which $\mathcal{D}^\ast$ decides that $(\bb A\not\indd\bb B| \bb C)_G$.

Note that the asynchrony of $\mc D^\ast$ stems from the fact that there exists no \emph{global clock} for the system and hence any node, upon receiving a message, follows Steps (i) and (ii) \emph{autonomously}, i.e., informally, without having to attend to what computations other nodes in $G$ are performing.

Some of the computational properties of the proposed algorithm $\mc D^\ast$ are formally articulated in Proposition~1 below.

\textbf{Proposition 1.} \emph{The following statements hold for $\mc D^\ast$.
\begin{itemize}
\item[(1)] For a given $d$-separation query $(\bb A \indd \bb B|\bb C)_G$ and DAG $G$, 
\begin{eqnarray*}
&\text{``$\bb C$ does not $d$-separate $\bb A$ from $\bb B$ in $G$"}\Longleftrightarrow \\
&\text{``Clash takes place during $\mc D^\ast$'s execution"}.
\end{eqnarray*}
\item[(2)] $\mc D^\ast$'s message-passing is confined within the ancestral graph $G_{An(\bb A\cup\bb B\cup\bb C)}$.
\item[(3)] During $\mc D^\ast$'s execution, either a clash between colors red ($\rb$) and green ($\gb$) takes place (see Remark~1) upon which $\mc D^\ast$ decides that $(\bb A\not\indd \bb B|\bb C)$, or a state of equilibrium will be reached in $O(l_{An(\bb A\cup\bb B\cup\bb C)})$ time where $l_{An(\bb A\cup\bb B\cup\bb C)}$ denotes the length of the longest undirected path in the ancestral graph $G_{An(\bb A\cup\bb B\cup\bb C)}$.
\item[(4)] Message-passing terminates in $O(1)$ time after reaching the state of equilibrium, thereby guaranteeing the termination of $\mc D^\ast$.
\item[(5)] Message-complexity of $\mc D^\ast$ is $O(|E_{An(\bb A\cup\bb B\cup\bb C)}|)$ where $E_{An(\bb A\cup\bb B\cup\bb C)}$ is the set of the edges of the ancestral graph $G_{An(\bb A\cup\bb B\cup\bb C)}$.  
\item[(6)] Communication-complexity of $\mc D^\ast$ is $O(|E_{An(\bb A\cup\bb B\cup\bb C)}|)$ bits where $E_{An(\bb A\cup\bb B\cup\bb C)}$ is the set of the edges of the ancestral graph $G_{An(\bb A\cup\bb B\cup\bb C)}$. 
\end{itemize}}

The reader is referred to Sec.~C-VI of Appendix C for the proof of Proposition~1.

\subsection{High-Level Understanding of $\mc D^\ast$}
\label{sec_high_level}
$\mc D^\ast$ has a simple machinery as we informally discuss here. Upon variables in $\bb A\cup\bb B\cup \bb C$ sending their colors to their parents, colors white ($\wb$), green ($\gb$), and red ($\rb$) begin to propagate in a \emph{backwards} manner throughout the network. In the midst of this process, white-color nodes which have a neighboring node colored either red ($\rb$) or green ($\gb$), change their color to that of their neighbors, and if a clash ever occurs between colors red and green, $\mc D^\ast$ decides that the input $d$-separation query is false (i.e., it is a \textsc{no}-instance $d$-separation query). \tb{Informally put, white-color nodes function as relays, which, by copying the colors of their neighbors, facilitate the possibility of a (permissible) collision between colors red and green.}

\subsection{A Note On The Termination of $\mc D^\ast$}
According to Proposition~1, if the input $d$-separation query presented to $\mathcal{D}^\ast$ is true (i.e., it is a \textsc{yes}-instance $d$-separation query), the system reaches a state of equilibrium in $O(l_{An(\bb A\cup\bb B\cup\bb C)})$ time and message-passing is guaranteed to terminate in $O(1)$ time after that. However, due to its local view, a node cannot know if such a global state has been reached. This is a fairly standard situation for an asynchronous distributed algorithm to find itself in \citep{mattern1987algorithms,tel2000introduction}, leading to the introduction of the fundamental concept of Termination-Detection (TD) in the distributed systems literature; see \citep[Ch.~8]{tel2000introduction}. There exist a variety of TD algorithms in the literature (e.g., \citealp{dijkstra1983derivation,mattern1987algorithms,mittal2004message,mittal2007family}). For example, \citet{mittal2004message} proposed two TD algorithms, each having detection latency of $O(D)$ where $D$ is the diameter of the underlying graph $G$, and $G$ is allowed to have an arbitrary topology.

\section{$\mc D^\ast$ in Action: A Case Study}
\label{sec_case_study}
\begin{figure*}[h!]
\centering
\includegraphics[width=0.77\textwidth]{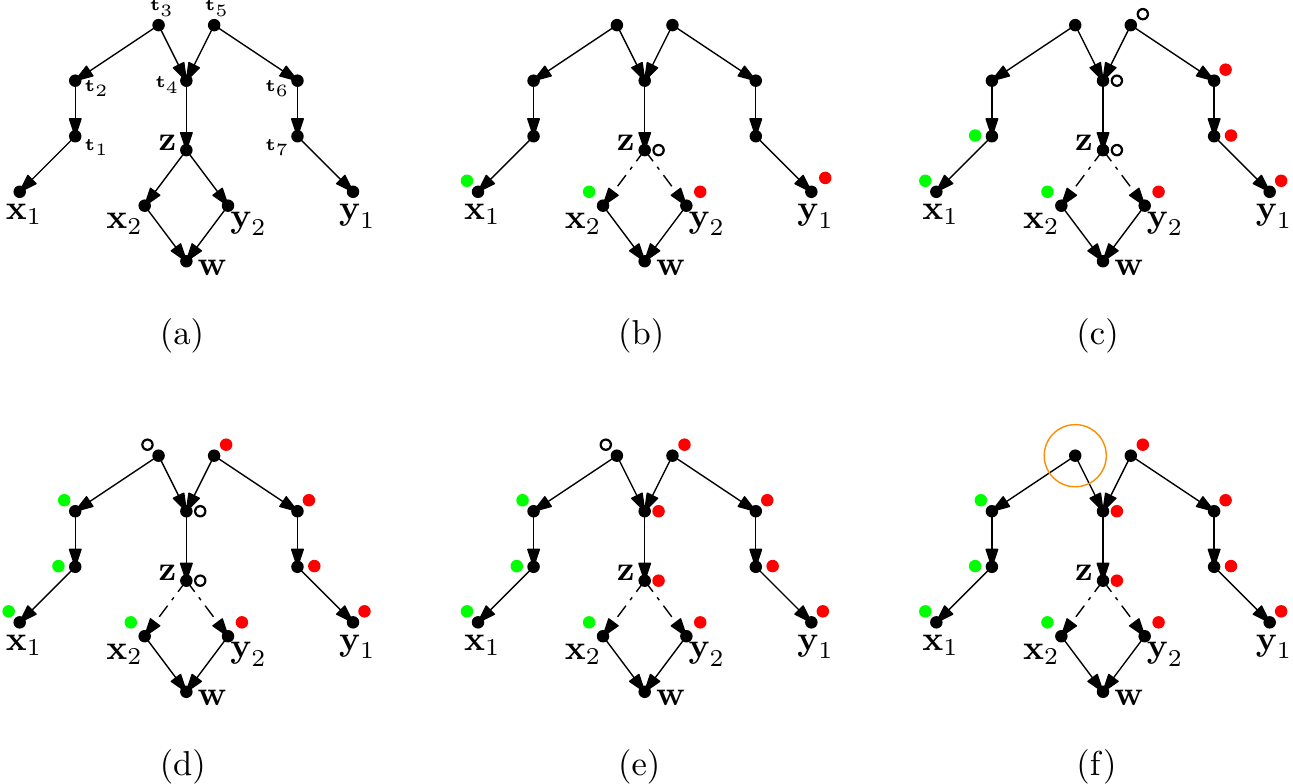}
\caption{\small{Illustrative example. The underlying DAG $G$ is shown in (a). The initial configuration of the system is portrayed in (b), wherein variables in sets $\bb X,\bb Y,$ $\bb Z$ self-activate in the states represented by green ($\gb$), red ($\rb$), and white ($\wb$), respectively. Depicting the downlinks of the variables in $\bb Z$ in a dash-dotted format simply symbolizes that the variables in $\bb Z$ ignore any message received from any of their children, and also do not send any message to any of their children.  $\mc D^\ast$ begins by nodes in $\bb X, \bb Y$, $\bb Z$ sending their colors as messages to their parents and proceeds as shown in (c-f) with each figure depicting a snapshot of the global state of the system at some instance in global time. Eventually, upon occurrence of a clash between colors green and red (at the circled node in (f)), $\mc D^\ast$ decides that $(\bb X\not\indd \bb Y|\bb Z)_G$.}}
\label{fig_motive}
\end{figure*}

In this section, we present an example to illustrate an execution and highlight the simplicity of $\mc D^\ast$. Consider the CBN depicted in Fig.~\ref{fig_motive}(a). Let the posed $d$-separation query be $(\bb X\indd \bb Y|\bb Z)_G$ where $\bb X=\{\bb x_1, \bb x_2\}$, $\bb Y=\{\bb y_1, \bb y_2\}$, and $\bb Z=\{\bb z\}$. According to the $d$-separation criterion \citep{pearl1988probabilistic}, observation of $\bb z$ activates the path $\bb x_1\leftarrow \bb t_1 \leftarrow \bb t_2 \leftarrow \bb t_3 \rightarrow \bb t_4 \leftarrow \bb t_5 \rightarrow \bb t_6 \rightarrow\bb t_7 \rightarrow \bb y_1$, thereby yielding the falsity of the $d$-separation query $(\bb X\indd \bb Y|\bb Z)_G$ (hence, the input is a \textsc{no}-instance query); see Fig.~\ref{fig_motive}(a). \tb{An execution of $\mc D^\ast$ is illustrated using successive \emph{snapshots} shown in Figs.~\ref{fig_motive}(b-f) with each figure depicting the global state of the system (i.e., nodes' colors) at some instance in global time (aka system's \emph{configuration}).}\footnote{\tbbb{Cast into Lamport's \emph{space-time diagram}  \citep{lamport1978time}, each figure depicts the global state of the system which corresponds to a vertical \emph{time-cut} positioned at a global time (see Mattern, 1987) and the time-cuts corresponding to Figs.~\ref{fig_motive}(b-f) are successively ordered.}} As depicted in Fig.~\ref{fig_motive}(b), variables in sets $\bb X,\bb Y,$ and $\bb Z$ initially self-activate in the states represented by colors green ($\gb$), red ($\rb$), and white ($\wb$), respectively. Also recall that, as explicated in Sec.~\ref{sec_alg_main}, variables in $\bb Z$ ignore any message received from any of their children, and also do not send any message to any of their children---depicting the downlinks of the variables in $\bb Z$ in a dash-dotted format simply  illustrates this statement pictorially in Fig.~\ref{fig_motive}(b). The colors green ($\gb$), red ($\rb$), and white ($\wb$) propagate in a backwards manner (Figs.~\ref{fig_motive}(c-d)). Also, the color of a white node gets replaced by green or red once a neighboring node acquires such colors (Figs.~\ref{fig_motive}(d-f)). Eventually, in the configuration depicted in Fig.~\ref{fig_motive}(f), a clash takes place between colors green and red at a node (circled node in Fig.~\ref{fig_motive}(f)), upon which $\mc D^\ast$ decides that $(\bb X\not\indd \bb Y|\bb Z)_G$.

Notice that, since $\bb w$ is unobserved (Fig.~\ref{fig_motive}(a)), the path $\bb x_2\rightarrow\bb w\leftarrow\bb y_2$ indeed remains blocked \citep{pearl2000causality}; this is nicely captured by the machinery of $\mc D^\ast$. Algorithm $\mc D^\ast$ prevents $\bb x_2$ and $\bb y_2$ from sending their colors in the forward direction (i.e., along the edges pointing to $\bb w$), thereby guaranteeing the occurrence of no clash along the blocked path $\bb x_2\rightarrow\bb w\leftarrow\bb y_2$. Also notice that, since $\bb z$ is observed (Fig.~\ref{fig_motive}(a)), the path $\bb x_2 \leftarrow \bb z \rightarrow \bb y_2$ is blocked as well \citep{pearl2000causality}. Once again the machinery of $\mc D^\ast$, due to $\bb z$ refraining from engaging in message-exchange with its children, ensures that no clash takes place due to the blocked path $\bb x_2 \leftarrow \bb z \rightarrow \bb y_2$.   

\section{Technical Discussion}
\label{sec_discussion}
A number of algorithms for the implementation of $d$-separation are proposed in the literature (\citealp{geiger1989d,lauritzen1990independence,shachter1998bayes,koller2009probabilistic,butz2016relevant}). Assuming $|E|\geq|V|$, to decide if $(\bb A\indd\bb B|\bb C)_G$ holds in $G$, the worst-case running time of Geiger et al.'s, Koller and Friedman's, Shachter's, and Butz et al.'s is $O(|E|)$ and that of Lauritzen et al.'s algorithm\footnote{The reader is referred to \citep{geiger1989d} for a detailed analysis of the running-time of Lauritzen et al.'s algorithm.} is $O(|V|^2)$ where $|V|$ and $|E|$ denote the number of the nodes and the edges of the underling DAG $G$, respectively. \tb{Note that, since for any DAG $G$, $|E|\leq |V|^2$, an $O(|E|)$-time algorithm (e.g., Geiger et al.'s) outperforms an $O(|V|^2)$-time algorithm (e.g., Lauritzen et al.'s) in terms of worst-case runtime\footnote{The gain in particularly significant in sparse graphs, where $|E|=O(|V|)$.} (see \citep{geiger1989d} for more discussions on this).} According to Proposition~1, the time-complexity of the proposed algorithm $\mc D^\ast$ is $O(l_{An(\bb A\cup\bb B\cup\bb C)})$ where $l_{An(\bb A\cup\bb B\cup\bb C)}$ denotes the length of the longest undirected path in the ancestral graph $G_{An(\bb A\cup\bb B\cup\bb C)}$. Since, for any DAG $G$, $l_{An(\bb A\cup\bb B\cup\bb C)}\leq |E|\leq |V|^2$, the proposed algorithm $\mc D^\ast$ outperforms all the previously proposed algorithms in terms of the worst-case running time.\btxt{\footnote{\btxt{According to Proposition~1, a \textsc{no}-instance $d$-separation query can be decided by $\mc D^\ast$ in time $O(l_{An(\bb A\cup\bb B\cup\bb C)})$; see also Sec.~C-II of the Appendix C. The upper-bound $O(l_{An(\bb A\cup\bb B\cup\bb C)})$ is an improvement over the worst-case runtime of all the previously proposed algorithms. Also note that, adopting a TD-algorithm with detection latency of $O(D)$ (see \citealt{mittal2004message,mittal2007family}, for such TD-algorithms), a \textsc{yes}-instance $d$-separation query can be decided by $\mc D^\ast$ in time $O(l_{An(\bb A\cup\bb B\cup\bb C)}+D)$ where $D$ is the diameter of $G$. Once again, since $l_{An(\bb A\cup\bb B\cup\bb C)}\leq |E|, D\leq |E|, |E|\leq |V|^2$, the upper-bound $O(l_{An(\bb A\cup\bb B\cup\bb C)}+D)$ is an improvement over the worst-case runtime of all the previously proposed algorithms. (Notice that, for any DAG $G$, $\frac{1}{2}(l_{An(\bb A\cup\bb B\cup\bb C)}+D)\leq |E|$, hence follows $|E|=\Omega(l_{An(\bb A\cup\bb B\cup\bb C)}+D)$.)}}} Particularly, the gain is significant in dense DAGs. Note that, in the limit as the underlying DAG $G$ gets denser, the worst-case runtime performances of the previously proposed algorithms become identical, i.e., $O(|V|^2)$.

The proposed algorithm $\mathcal{D}^\ast$ restricts its exploration solely in the ancestral graph\linebreak $G_{An(\bb A\cup\bb B\cup\bb C)}$, as formalized by Statement (2) of Proposition~1. The idea of exploring the ancestral graph is at the core of Lauritzen et al.'s algorithm for $d$-separation \citep{lauritzen1990independence}. However, in sharp contrast to Lauritzen et al.'s algorithm, the proposed algorithm $\mc D^\ast$ need not \emph{moralize} the ancestral graph $G_{An(\bb A\cup\bb B\cup\bb C)}$. As \citet{geiger1989d} point out, the moralization step of Lauritzen et al.'s algorithm requires $O(|V|^2)$ time in the worst-case. 

Another noteworthy property of $\mathcal{D}^\ast$ is its tendency toward quick detection of false $d$-separation queries (i.e., \textsc{no}-instance queries), manifested in an occurrence of a clash according to Remark~1. For a \textsc{no}-instance $d$-separation query, Proposition~2, below, gives a more refined upper-bound on the time required for an occurrence of a clash, thereby formalizing the said claim. The reader is referred to \tb{Sec.~C-III} of Appendix C for the proof of Proposition~2.

\textbf{Proposition~2.} \emph{Let $\displaystyle\bb A=\{\bb a_i\}_i$, $\bb B=\{\bb b_j\}_j$, $\bb C=\{\bb c_k\}_k$ be three disjoint sets of nodes belonging to a DAG $G$. Let $l_{An(\bb A\cup\bb B\cup\bb C)}^d$ denote the length of the longest directed path in the ancestral graph $G_{An(\bb A\cup\bb B\cup\bb C)}$, and   $l_{An(\bb A\cup\bb B\cup\bb C)}^{ij}$ the length of the shortest unblocked path between the nodes $\bb a_i$ and $\bb b_j$ in $G_{An(\bb A\cup\bb B\cup\bb C)}$. As a convention, if all paths between $\bb a_i$ and $\bb b_j$ are blocked, $l_{An(\bb A\cup\bb B\cup\bb C)}^{ij}=\infty$. If $(\bb A \not\indd \bb B| \bb C)_G$ then a clash between colors green \emph{($\gb$)} and red \emph{($\rb$)} occurs in time $\displaystyle O\big( l_{An(\bb A\cup\bb B\cup\bb C)}^d+\min_{i,j} l_{An(\bb A\cup\bb B\cup\bb C)}^{ij}\big)$, upon which $\mc D^\ast$ decides that $(\bb A \not\indd \bb B| \bb C)_G$.}

In Sec.~\ref{sec_notation}, we formally defined a notion called refutation-module (see Def.~1). In the language of computational complexity and theorem-proving, a refutation-module $\mc M_{(\bb X \not\indd \bb Y|\bb Z)_G}$ can serve as a \emph{certificate} (or \emph{witness}) for disproving a $d$-separation query $(\bb X \indd \bb Y|\bb Z)_G$. This interpretation is related to the verifier-based definition of the complexity class \textit{coNP}. Next, in Proposition~3, we provide an even more refined upper-bound on the time required for an occurrence of a clash, thereby strengthening our claim as to $\mathcal{D}^\ast$'s  tendency toward quick detection of false $d$-separation queries. The reader is referred to \tb{Sec.~C-IV} of Appendix C for the proof of Proposition~3.

\textbf{Proposition~3.} \emph{Let $\bb X,\bb Y,\bb Z$ be three disjoint sets of nodes belonging to a DAG $G$. Also, let $(\bb X \not\indd \bb Y|\bb Z)_G$. Let $\mc M_{(\bb X \not\indd \bb Y|\bb Z)_G}$ denote a refutation-module for the \rtxt{query} $(\bb X \indd \bb Y|\bb Z)_G$ with $l_{\mc M}^d$ and $|P_{\mc M}|$ denoting the length of the longest directed path and the shortest unblocked path in $\mc M_{(\bb X \not\inddd \bb Y|\bb Z)_G}$, respectively. \tb{Finally}, let $\mc M_{(\bb X \not\inddd \bb Y|\bb Z)_G}^{\ast}$ denote the minimal refutation-module for the \rtxt{query} $(\bb X \indd \bb Y|\bb Z)_G$, with $E_{\mc M_{(\bb X \not\inddd  \bb Y|\bb Z)_G}^{\ast}}$ denoting the set of the edges of $\mc M_{(\bb X \not\inddd \bb Y|\bb Z)_G}^{\ast}$. Then the following statement holds true: A clash between colors green \emph{($\gb$)} and red \emph{($\rb$)} occurs in time $O(\min_{\mc M_{(\bb X \not\indd \bb Y|\bb Z)_G}}\{l_{\mc M}^d+|P_{\mc M}|\})\leq O(|E_{\mc M_{(\bb X \not\indd \bb Y|\bb Z)_G}^{\ast}}|)$, upon which $\mc D^\ast$ decides that $(\bb X \not\indd \bb Y| \bb Z)_G$.}

To further highlight the significance of Proposition~3, let us consider the following \emph{nondeterministic} algorithm $\mc A$. Algorithm $\mc A$ takes as input a DAG $G$ along with a $d$-separation query $(\bb X \indd \bb Y|\bb Z)_G$, and outputs \textsc{yes} or \textsc{no} depending on whether the input query is a \textsc{yes}-instance or a \textsc{no}-instance query, respectively.

\begin{itemize}
\item[(i)] Nondeterministically guess (1) the minimal refutation-module $\mc M_{(\bb X \not\inddd \bb Y|\bb Z)_G}^{\ast}$ in $G$ for the $d$-separation \rtxt{query}\linebreak $(\bb X \indd \bb Y|\bb Z)_G$ (by definition, $\mc M_{(\bb X \not\inddd \bb Y|\bb Z)_G}^{\ast}$ contains an active path, $P^{\ast}$, between a node $\bb x^\ast\in \bb X$ and a node $\bb y^\ast \in \bb Y$, and also contains a set of observed variables $\bb Z^\ast\subseteq \bb Z$)\footnote{Note that if $P^\ast$ does not contain any head-to-head node, then $\bb Z^\ast=\varnothing$}, and (2) the corresponding nodes $\bb x^\ast,\bb y^\ast,\bb Z^\ast$ belonging to $\mc M_{(\bb X \not\indd \bb Y|\bb Z)_G}^{\ast}$.

\item[(ii)] Verify that (1) $\bb x^\ast \in \bb X$, $\bb y^\ast \in \bb Y$, and $\bb Z^\ast\subseteq \bb Z$ (this can be straightforwardly verified in $O(|\bb X|+|\bb Y|+|\bb Z^\ast||\bb Z|)$ time), (2) $\mc M_{(\bb X \not\indd \bb Y|\bb Z)_G}^{\ast}$ is a subgraph of $G$ (this can be straightforwardly verified in $O(|E_{\mc M_{(\bb X \not\inddd \bb Y|\bb Z)_G}^{\ast}}|)$ time), and (3) $d$-separation relation $(\bb x^\ast\indd \bb y^\ast|\bb Z^\ast)$ does not hold in DAG $\mc M_{(\bb X \not\inddd \bb Y|\bb Z)_G}^{\ast}$ (this can be verified in $O(|E_{\mc M_{(\bb X \not\inddd  \bb Y|\bb Z)_G}^{\ast}}|)$ time, using Geiger et al.'s algorithm \citep{geiger1989d}). If all the verification steps (1)-(3) pass, output \textsc{no}; otherwise, output \textsc{yes}.
\end{itemize}

Altogether, presented with a \textsc{no}-instance $d$-separation query $(\bb X \indd \bb Y|\bb Z)_G$, algorithm $\mc A$ outputs \textsc{no} in $O(|E_{\mc M_{(\bb X \not\indd \bb Y|\bb Z)_G}^{\ast}}|+|\bb X|+|\bb Y|+|\bb Z^\ast||\bb Z|)$ nondeterministic time. Interestingly according to Proposition~3, presented with a \textsc{no}-instance $d$-separation query $(\bb X \indd \bb Y|\bb Z)_G$, the machinery of $\mc D^\ast$ ensures that a clash between colors green ($\gb$) and red ($\rb$) occurs within at most $O(|E_{\mc M_{(\bb X \not\indd \bb Y|\bb Z)_G}^{\ast}}|)$ time, upon which $\mc D^\ast$ decides that $(\bb X \not\indd \bb Y| \bb Z)_G$. It is crucial to note that the presented argument solely concerns \textsc{no}-instance $d$-separation queries.

Proposition~4, given below, further strengthens the claim of Proposition~3. The reader is referred to \tb{Sec.~C-V} of Appendix C for the proof of Proposition~4.

\textbf{Proposition~4.} \emph{The upper-bound $O(\min_{\mc M_{(\bb X \not\indd \bb Y|\bb Z)_G}}\{l_{\mc M}^d+|P_{\mc M}|\})$ given in Proposition~3 is tighter than the one given in Proposition~2.}

Finally, we would like to point out an interesting property of the CUG, referred to as \emph{order-invariance}, which is characterized informally as follows: \emph{The order according to which nodes in the network receive their messages is irrelevant.} More formally, the order-invariance property can be stated as follows: Assume that a node $\bb x$ is at state $S_{i}^{\bb x}$ and upon receiving the sequence of messages $M_1, M_2, \cdots, M_n$ ends up in state $S_{f}^{\bb x}$. Then the following holds true for the node $\bb x$. For any permutation $\pi$ defined on the set $\{1,2,\cdots,n\}$, the node $\bb x$, starting at state $S_{i}^{\bb x}$, would end up in the state $S_{f}^{\bb x}$ upon receiving the sequence of messages $M_{\pi(1)}, M_{\pi(2)}, \cdots, M_{\pi(n)}$. The reader is referred to \tb{Sec.~C-VIII} of Appendix C for a formal treatment of the order-invariance property and its proof.

\section{General Discussion}
\label{d_sep_sec_neuro_psych}
The Algorithm $\mc D^\ast$, in the spirit of Pearl's \citeyearpar{pearl1986fusion} belief propagation scheme, employs the edges of the underlying CBN as the medium through which message-passing between nodes takes place. The latter echos Pearl's \citeyearpar{pearl1986fusion} insight when he advocated the idea that a CBN must \emph{not} be viewed {``merely as a passive parsimonious code for storing factual knowledge but also as a computational architecture for reasoning about that knowledge."} $\mc D^\ast$ adheres to this idea. Recent literature in neuroscience investigating possible implementation of CBNs at the neural level supports Pearl's idea (see  \citealp{lochmann2011neural,gershman2016complex}). Lochmann and Deneve \citeyearpar{lochmann2011neural} advocate the idea that a CBN's node can be represented at the neural level by a single \citep{deneve2008bb2,deneve2008bb1} or a population of neurons \citep{ma2006bayesian} with the neural network resembling a  ``mirror image" of the CBN it implements---though sometimes not a `perfect' mirror---and the links of the neural network providing the medium for inference to be carried out---either in the form of belief propagation or sample-based methods like Gibbs sampling.\footnote{For more on how probability distributions can be encoded at the neural level, the reader is referred to Lochmann and Deneve \citeyearpar{lochmann2011neural}.}

Interestingly, the peculiar tendency of $\mc D^\ast$ toward quick detection of \textsc{no}-instance $d$-separation queries is consistent with our pre-theoretical intuition that humans tend to detect possible dependencies between concepts and propositions rather swiftly, once such dependencies do exist. The following question then presents itself: Could this tendency be supported based on any rational grounds? In what follows we provide an argument supporting the rationality of the foregoing tendency. $(\dagger)$ Assuming that the mind incurs a higher rate of loss (defined as incurred cost per unit of time) for discovering a dependency when one does exist, compared to the condition wherein one does not exist and the mind recognizes that, we formally show that the foregoing tendency is simply a consequence of the mind acting as a boundedly-rational {satisficer} \citep{simon1957models}, trying to attain good performance in terms of {expected} runtime (i.e., average-case analysis). But why should the rate of loss under the condition wherein a dependency does exist be higher? Informally put, why should the mind be so hasty in detecting dependencies under that condition? One possible explanation is that it is crucial for the mind to swiftly detect dependencies under that condition, with the rationale being that delay in detecting those dependencies could be harmful to the reasoner and potentially jeopardize their life, hence important from an evolutionary standpoint. Furthermore, given the prominent role that explanation and inference play in human cognition {(see \citealp{lombrozo2016explanatory})}, it is crucial for the mind to promptly detect those factors deemed (probabilistically) relevant to the task faced by the reasoner.

Let us formally characterize a general condition under which the aforesaid tendency can be given a rational basis. Let $\bb T_{\mc A}$ denote the runtime of an algorithm $\mc A$ implementing $d$-separation criterion, $\pi_{\textsc{yes}}$ and $\pi_{\textsc{no}}$ denote the prior probability of the input being a \textsc{yes}-instance and \textsc{no}-instance $d$-separation query, respectively. Let also $\bb T_{\mc A}^{\textsc{yes}}$ and $\bb T_{\mc A}^{\textsc{no}}$ denote the worst-case runtime of $\mc A$ on \textsc{yes}-instance and \textsc{no}-instance $d$-separation queries, respectively. Finally, let $\mc L_{\textsc{yes}}$ and $\mc L_{\textsc{no}}$ denote the cost per unit of time incurred by $\mc A$ for delay in detecting a \textsc{yes}-instance and \textsc{no}-instance $d$-separation query, respectively. Then, for any underlying DAG $G$, the following holds true: $\mathbb{E}[\bb T_{\mc A}]\leq \frac{\mc L_{\textsc{yes}}}{\mc L_{\textsc{yes}}+\mc L_{\textsc{no}}}\bb T_{\mc A}^{\textsc{yes}}\pi_{\textsc{yes}}+\frac{\mc L_{\textsc{no}}}{\mc L_{\textsc{yes}}+\mc L_{\textsc{no}}}\bb T_{\mc A}^{\textsc{no}}\pi_{\textsc{no}},$ where the expectation $\mathbb{E}[\cdot]$ is taken with respect to the (unknown) distribution of all $d$-separation queries. It is then easy to show that, under the condition $(\ast)\ \mc L_{\textsc{no}}\pi_{\textsc{no}}\geq \mc L_{\textsc{yes}}\pi_{\textsc{yes}},$ it is rational for the mind trying to attain good performance in terms of {expected} runtime to demonstrate the said tendency toward quick detection of \textsc{no}-instance $d$-separation queries. The setting portrayed in $(\dagger)$ above is a special case of Condition~$(\ast)$: It corresponds to Condition~$(\ast)$ subject to the assumptions $\pi_{\textsc{no}}=\pi_{\textsc{yes}}$ (reflecting the reasoner's uninformative, \emph{a priori} expectation that \textsc{yes}- and \textsc{no}-instance queries are equiprobable) and $\mc L_{\textsc{no}}\geq\mc L_{\textsc{yes}}$ (reflecting a higher rate of loss for erring on \textsc{no}-instance queries, as alluded to earlier). Future work should experimentally investigate if humans demonstrate the forgoing normatively justified tendency in probabilistic (in)dependence judgment tasks, or that, on the contrary, they systematically deviate from that.

Also interestingly, the forgoing tendency of $\mc D^\ast$ toward focusing its search on the minimal refutation module can be taken as evidence for its least-effort-like characteristic, and is fully consistent with recently proposed frameworks which seek rational understanding of the mind at the algorithmic level of analysis by appealing to the notion of economical use of limited computational and cognitive resources (in our case, by striving for minimizing the size of the module required to be investigated for refuting a false $d$-separation query); see {Nobandegani (2017) and Griffiths et al.~(2015)}. Although we briefly discussed the idea of termination detection for asynchronous distributed algorithms, a boundedly-rational agent may decide to only run an asynchronous distributed algorithm for a period of time which is justified based on the opportunity cost incurred by delaying another task. In that light, the boundedly-rational agent may plausibly decide to adopt termination detection algorithms only in settings wherein the opportunity costs involved would be relatively low. Also notably, $\mc D^\ast$  exemplifies how the pursuit of cognitive plausibility can lead to the discovery of state-of-the-art algorithms.

Perhaps the biggest limitation of $\mc D^\ast$ (and, likewise, of belief propagation) is the assumption that communication channels are faultless, allowing for reliable message exchange. The brain's neural circuits involve much stochasticity and response variability \tbbb{(e.g., Ma \& Jazayeri, 2014; Ma, Beck, and Pouget, 2008; Summerfield \& Tsetsos, 2015)}, undermining this assumption. Future work should investigate extensions of $\mc D^\ast$ that are more robust to neural noise. While many questions remain open, we hope to have made some progress toward understanding human probabilistic (in)dependence judgment at the algorithmic level, a capacity without which the world would seem too chaotic for humans to live by.

\vspace*{5pt}
\hspace*{-10pt}\textbf{Acknowledgments:} {We would like to  thank Michael Pacer for providing constructive comments on an earlier draft of this work, and, Tom Shultz and Luc Devroye for helpful discussions. This work was supported in part by the Natural Sciences and Engineering
Research Council of Canada under grant RGPIN 262017.}

\section*{\centering \Large Appendix C}
Throughout Appendix C, let $(\bb A \indd \bb B|\bb C)_G$ denote the posed $d$-separation query with DAG $G$ representing the topology of the underlying BN. Throughout the proofs and arguments to follow, it is assumed that communication channels are reliable, bidirectional, and first-in first-out (FIFO) \citep{lynch1996distributed}. For the time-complexity analysis of $\mc D^\ast$, we adhere to the same assumptions adopted in \citep{lynch1996distributed}. More specifically, we assume: (ASM-1) an upper-bound of $\alpha$ for a process to perform Steps (i) and (ii) upon receipt of a message, and  (ASM-2) an upper-bound of $\beta$ on the delivery time for each message in a channel. Note that the parameters $\alpha$ and $\beta$ are arbitrary but finite constants. Also note that, as the number of messages exchanged by $\mc D^\ast$ on an edge is $O(1)$ (see Statement (5) of Proposition~1 in the main text), the effect of pileups (aka congestion) on a channel has been considered in Assumptions (ASM-1) and (ASM-2).\footnote{Since the number of messages exchanged by $\mc D^\ast$ on an edge is $O(1)$, the following holds: (a) the number of messages in any channel queue is at most $O(1)$, and (b) the number of messages awaiting in a process's send buffer is at most $O(1)$.}

\section*{C-I\hspace*{10pt} $\mc D^\ast$: Proof of Correctness}
\label{Alg_proof_correctness} 
In what follows, we prove three statements which, taken together, grant the correctness of $\mc D^\ast$. The three statements are given below.

\begin{itemize}
\item[(I)] For a given $d$-separation query $(\bb A \indd \bb B|\bb C)_G$ and DAG $G$, 
\begin{eqnarray*}
&\text{``$\bb C$ does not $d$-separate $\bb A$ from $\bb B$ in $G$"}\Longleftrightarrow &\\ 
&\text{``Clash takes place during $\mc D^\ast$'s execution"}.&
\end{eqnarray*}
\item[(II)] During $\mc D^\ast$'s execution, either a clash between colors red ($\rb$) and green ($\gb$) takes place (cf. Remark~1 in the main text) upon which $\mc D^\ast$ decides that $(\bb A\not\perp \bb B|\bb C)$, or a state of equilibrium will be eventually reached.
\item[(III)] Message-passing terminates in $O(1)$ time after reaching the state of equilibrium, thereby guaranteeing the termination of $\mc D^\ast$.
\end{itemize} 

The proofs of Statements (I) to (III) are presented next.

\subsection*{C-I.I \hspace*{10pt}Proof of Statement (I)}
\label{sec_s_I_I}
Next, the proof of Statement (I) is presented. First, the proof of the forward direction is outlined ({Sec.~C-I.I.I}), followed by the proof of the backward direction ({Sec.~C-I.I.II}).

\subsubsection*{C-I.I.I \hspace*{10pt}Proof of Statement (I): Forward Direction}
\label{state_1_forward_proof}
\begin{figure}[h!]
\centering
\includegraphics[width=0.32\textwidth]{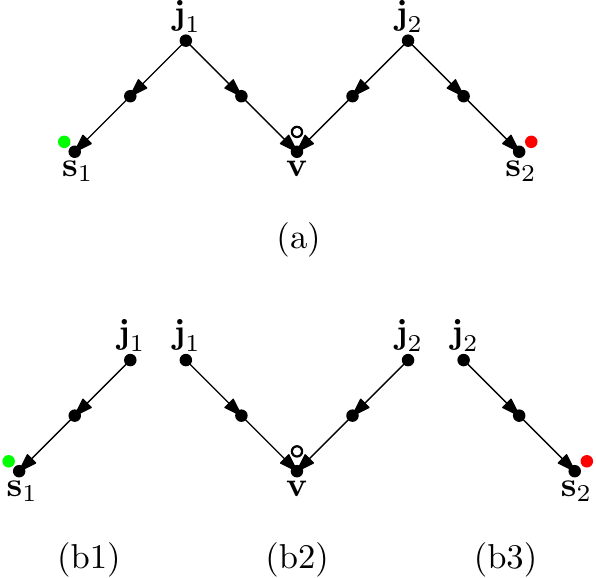}
\caption{Decomposition of a generic unblocked path into v-structured and non-v-structured modules. \textbf{(a)} A generic unblocked path $p$ comprised of v-structured as well as non-v-structured modules. The nodes $\bb s_1$ and $\bb s_2$ are {source} vertices. The nodes $\bb j_1$ and $\bb j_2$ are {joint} vertices. The node $\bb v$ is a collider. Without loss of generality, $\bb s_1, \bb s_2$, and $\bb v$ are assumed to be initialized with colors green, red and white, respectively. \textbf{(b1)} A non-v-structured module of the unblocked path $p$ with the source vertex $\bb s_1$. \textbf{(b2)} The v-structured module of the unblocked path $p$. \textbf{(b3)} A non-v-structured module of the unblocked path $p$ with the source vertex $\bb s_2$.}
\label{fig_proof_f}
\end{figure}

We prove the forward direction of Statement (I) next. This is accomplished by proving the following: Conditioned on the set $\bb C$, if there exists an unblocked path between $\bb a\in \bb A$ and $\bb b\in\bb B$ (for any $\bb a,\bb b$), a clash of the kind stated in Remark 1 is unavoidable during $\mc D^\ast$'s execution. A path $l$ is said to be unblocked \citep{pearl1988probabilistic} if and only if (a) for every collider node $\bb n$ on $l$, either $\bb n$ or some of $\bb n$'s descendants are in $\bb C$, and (b) for every non-collider node $\bb m$, $\bb m\not\in \bb C$. The proof rests on a simple understanding that a generic unblocked path can be decomposed into v-structured and non-v-structured \emph{modules} as illustrated in Fig.~\ref{fig_proof_f}. Neighboring modules share a common vertex which we refer to as \emph{joint vertex} (e.g., the nodes $\bb j_1,\bb j_2$ in Fig.~\ref{fig_proof_f}(a)). The end-point vertex of a non-v-structured subpath which is not a joint vertex is termed \emph{source vertex}; see Fig.~\ref{fig_proof_f}(b1) and Fig.~\ref{fig_proof_f}(b3).
In principle, an unblocked path may have multiple v-structured modules. For ease of exposition, the unblocked path $p$ depicted in Fig.~\ref{fig_proof_f}(a) possesses only one v-structured module. Note that the proof that follows does not make this restrictive assumption.

Next, we prove the inevitability of a clash for unblocked paths possessing non-v-structured as well as v-structured modules.\footnote{The adaptation of the argument for the other two cases where the unblocked path is solely comprised of either non-v-structured modules or v-structured modules is straightforward.} The proof comprises two parts. In Part I, we show the inevitability of a clash over such an unblocked path, $l^\ast$, provided that no message is destined from a node outside $l^\ast$ to a node belonging to $l^\ast$. Using the arguments provided in Part I, in Part II we show that \emph{regardless} of the messages destined from nodes outside $l^\ast$ to the nodes belonging to $l^\ast$, the occurrence of a clash on $l^\ast$ is inevitable (i.e., eventually happens).

\textbf{Proof of Part I:} Rules $(\var,\gb)\rightarrow \gb$, and $(\var,\rb)\rightarrow \rb$ sketched in Step (ii) of $\mc D^\ast$, along with $\mc D^\ast$'s initialization phase wherein all the nodes in the sets $\bb A, \bb B,$ and $\bb C$ propagate their colors to their parents, ensure that all non-v-structured modules are fully explored and, by the end of exploration, all the nodes within each non-v-structured module will be homogeneously colored consistent with that of the respective source vertex, except for the joint vertex which requires more careful consideration ($\dagger$). The propagation of white ($\wb$) through the DAG $G$ in a backward manner ensures that v-structured modules are fully explored and, by the end of exploration, all the nodes within each v-structured module will be homogeneously colored in white ($\wb$), except for the joint vertices which require more careful consideration ($\ddagger$). The consideration advised in ($\dagger$) and ($\ddagger$) is explicated next.\footnote{The analysis of the case for the joint vertex between two adjacent v-structured modules does not require special consideration since it will become white ($\wb$) first and thereafter, according to the CUG, will function as a \emph{relay}, transferring the color of one branch to the other.} The joint vertex connecting a non-v-structured module to a v-structured module may first become white ($\wb$) or whatever the color of the source vertex of the non-v-structured module is, depending on whether the joint vertex first receives a message from the non-v-structured module or the v-structured module, respectively. However, and quite importantly, its color eventually becomes that of the source vertex of the non-v-structured module and, according to Step (ii) of $\mc D^\ast$, it sends its color down the v-structured module. In short, any joint vertex $\bb j$ will eventually serve as a \emph{relay} transferring the color of one side to the other in one of the following two ways: (1) either $\bb j$ becomes white and then, upon receiving a red- or green-colored message from a neighbor on one side, $\bb j$ changes it color and sends its new color down the other side, or (2) $\bb j$ first becomes green or red (due to receiving, respectively, a green or red message from a neighbor on one side) and then receives a white-colored message ($\wb$) from a neighbor $\bb w$ residing on the other side, upon which---in an act analogous to \emph{handshaking} in communication networks---$\bb j$ sends back its color to $\bb w$ which, in turn, initiates a chain reaction thereby $\bb w$ and its white-colored neighbors alter their color to that of $\bb j$ and so do their white-colored neighbors and so forth. This key understanding that joint vertices, as just explained, essentially serve as a relay transferring the color of one module to the other neighboring module, in addition to the fact that the color of the two source vertices are different, together, grants the conclusion that a clash between the colors green and red along the unblocked path $l^\ast$ eventually takes place. This concludes the proof of Part I.

\textbf{Proof of Part II:} As stated earlier, Part II concerns with showing the following: A message received by a node belonging to an unblocked path $l^\ast$ which is sent from a node lying outside $l^\ast$ cannot prevent the clash from happening on $l^\ast$. That is, informally, the occurrence of a clash cannot be prevented by any message coming from a node residing outside $l^\ast$ to a one belonging to $l^\ast$, say $\bb n_{\text{in}}$. We consider all the possible scenarios (i.e., scenarios (c1) to (c6) listed below) and show that indeed the claim of Part II holds true. Before we proceed further, let us introduce a notation. Let $\dbc{\alpha}{\beta}$ denote the following: $\bb n_{\text{in}}$'s current color is $\alpha$ and the color of the message (coming from a node residing outside $l^\ast$) destined to $\bb n_{\text{in}}$ is $\beta$. For example, $\dbc{\wb}{\rb}$ implies that $\bb n_{\text{in}}$'s current color is white and the incoming message is red.
\begin{itemize}
\item[(c1)] For $\dbc{\wb}{\wb}$, $\dbc{\gb}{\gb}$, $\dbc{\rb}{\rb}$: According to the CUG, if $\bb n_{\text{in}}$ receives a message whose color is identical to its current color, $\bb n_{\text{in}}$'s current color persists. Hence, the claim of Part II remains true under such circumstances.
\item[(c2)] For $\dbc{\rb}{\gb}$, $\dbc{\gb}{\rb}$: According to the CUG, these cases immediately lead to the occurrence of a clash. Hence, the claim of Part II remains valid under such circumstances.
\item[(c3)] For $\dbc{\gb}{\wb}$, $\dbc{\rb}{\wb}$: According to the CUG, if $\bb n_{\text{in}}$'s current color is green or red, it preserves its color upon receiving a white-colored message. Hence, the claim of Part II holds true under such circumstances.
\item[(c4)] For $\dbc{\varnothing}{\gb}$, $\dbc{\wb}{\gb}$: According to the CUG, if $\bb n_{\text{in}}$'s current color is white or $\bb n_{\text{in}}$ does not currently have any color, upon receiving a green-colored message, $\bb n_{\text{in}}$'s color becomes green and thereafter it acts as a green-colored source vertex for  $l^\ast$.\footnote{More specifically, once $\bb n_{\text{in}}$ becomes green it acts as a green-colored source vertex for the two \emph{subpaths} of $l^\ast$ which lie at the two sides of $\bb n_{\text{in}}$ and share $\bb n_{\text{in}}$ as their common node. For example, for the path $\bb v_1\leftarrow\bb v_2\leftarrow\bb v_3\rightarrow\bb n_{\text{in}}\rightarrow\bb v_4\leftarrow\bb v_5$, the two subpaths are $\bb v_1\leftarrow\bb v_2\leftarrow\bb v_3\rightarrow\bb n_{\text{in}}$ and $\bb n_{\text{in}}\rightarrow\bb v_4\leftarrow\bb v_5$.} This can only expedite the occurrence of a clash on $l^\ast$.  
\item[(c5)] For $\dbc{\varnothing}{\rb}$, $\dbc{\wb}{\rb}$: The line of reasoning is similar to the one given for (c4).
\item[(c6)] For $\dbc{\varnothing}{\wb}$: According to the CUG, this interaction changes $\bb n_{\text{in}}$'s color to white. However, due to the machinery of $\mc D^\ast$, color white merely acts as a placeholder awaiting to be replaced by green or red upon interacting with one of the kind. In this light, altering the state of $\bb n_{\text{in}}$ from $\varnothing$ to white cannot prevent a clash from happening on $l^\ast$.
\end{itemize}
This concludes the proof of Part II and, together with Part I, concludes the proof of the forward direction of Statement (I). 

\hfill $\blacksquare$
 
\subsubsection*{C-I.I.II \hspace*{10pt} Proof of Statement (I): Backward Direction}
\label{state_1_backward_proof}

\begin{figure}[h!]
\centering
\includegraphics[width=0.45\textwidth]{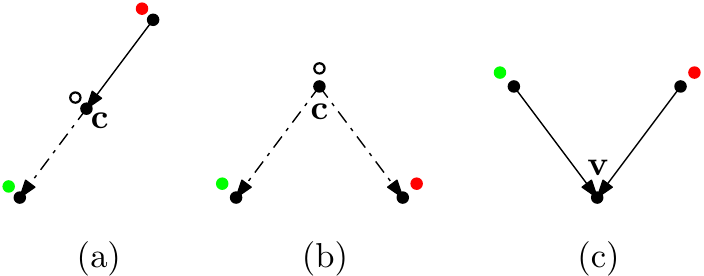}
\caption{The three types of subpaths. Depicting the downlinks of a variable $\bb c\in \bb C$ in a dash-dotted format simply symbolizes \tb{a crucial property of $\mc D^\ast$ according to which} $\bb c$ ignores any message received from any of its children, and also does not send any message to any of its children. \textbf{(a)} The green node and the red node are separated by a head-to-tail variable $\bb c\in \bb C$. \textbf{(b)} The green node and the red node are separated by a confounder $\bb c\in \bb C$. \textbf{(c)} The green node and the red node are separated by a collider $\bb v$ where neither $\bb v$ nor any of $\bb v$'s descendants is in the set $\bb C$.} 
\label{fig_proof_b}
\end{figure}

We prove the backward direction of Statement (I) next, using proof by contraposition. That is,  we prove: For a given $d$-separation query $(\bb A \indd \bb B|\bb C)_G$ and DAG $G$,
\begin{eqnarray*}
&\text{``$\bb C$ $d$-separates $\bb A$ from $\bb B$ in $G$"}\Rightarrow & \\
&\text{``Clash does not take place during $\mc D^\ast$'s execution"}.&
\end{eqnarray*}
According to \citep{pearl1988probabilistic}, the statement ``$\bb C$ $d$-separates $\bb A$ from $\bb B$ in $G$" is equivalent to the following: Every path between any $\bb a\in \bb A$ and any $\bb b\in \bb B$ is blocked. According to \citep{pearl1988probabilistic}, a path $l$ is said to be blocked if and only if at least one of the two statements holds: (a2) There exists a collider node $\bb n$ on $l$ where neither $n$ nor any of $\bb n$'s descendants is in $\bb C$, (b2) There exists a non-collider node $\bb m$ on $l$ where $\bb m\in \bb C$. Therefore, altogether, the statement ``$\bb C$ $d$-separates $\bb A$ from $\bb B$ in $G$" is equivalent to the statement that every path connecting $\bb a\in \bb A$ and $\bb b\in \bb B$ has to at least contain a \emph{subpath} of the type specified in (a2) and (b2). Hence, for a clash to take place on path $l$, one of the colors green or red has to pass through $l$'s corresponding subpath and collide with the other color. In what follows, we consider all such subpaths and show that, the very existence of such subpaths on every path connecting $\bb a\in \bb A$ and $\bb b\in \bb B$, grants the impossibility of an occurrence of a clash during $\mc D^\ast$'s execution. These subpaths can be of three types: (1) the green node and the red node are separated by a head-to-tail node which is observed (Fig.~\ref{fig_proof_b}(a)), (2) the green node and the red node are separated by a common cause (aka confounder) which is observed (Fig.~\ref{fig_proof_b}(b)), and finally (3) the green node and the red node are separated by a common effect (aka collider) which is neither itself nor any of its descendants is observed (Fig.~\ref{fig_proof_b}(c)). 

Next, we consider each case at a time and prove that $\mc D^\ast$'s machinery prevents the occurrence of a clash along any of the aforesaid subpaths depicted in Figs~\ref{fig_proof_b}(a-c). The proof for (1) and (2) immediately follows form the following crucial property of $\mc D^\ast$: \tb{Variables in $\bb C$ ignore any message received from any of their children, and also do not send any message to any of their children (depicting the outgoing edges from $\bb c$ in Figs~\ref{fig_proof_b}(a-b) simply symbolizes this property).} Case (3) requires more careful consideration. The only way for color green/red (on the one side) to reach color red/green (on the other side)---thereby generating a clash---was for the collider to be white-colored so that, by being replaced by either green or red, it would allow colors green and red to meet and hence a clash would occur. However, since (i) neither the collider nor any of its descendants is observed (and hence none of them are white), and also (ii) $\mc D^\ast$'s machinery dictates the propagation of the color white in a \emph{backwards} manner through the corresponding ancestors of the white-colored nodes, altogether, the collider cannot become white during an execution of $\mc D^\ast$. This concludes the proof. \hfill $\blacksquare$

\subsection*{C-I.II \hspace*{10pt}Proof of Statement (II)}
\label{appx_equlibrium_clash}
\begin{figure}[h!]
\centering
\includegraphics[width=0.23\textwidth]{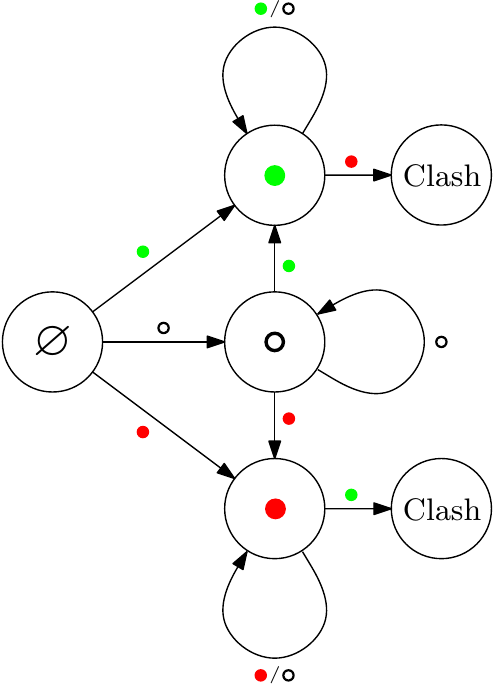}
\caption{State transition diagram. The message which ought to be received for a transition to take place is depicted on the corresponding edge. In case multiple messages engender the same transition, they are all detailed on the corresponding edge separated by slashes.}
\label{fig_transition}
\end{figure}

The state transition diagram for $\mc D^\ast$ is given in Fig.~\ref{fig_transition}. The states represent a node's color and the edges represent transitions due to receiving messages whose colors are depicted on the edges.

A simple inspection of the diagram reveals that a node's color cannot alternate between any two states. This is due to the fact that the diagram has no cycles of length two or greater. This observation implies that either a clash takes place upon which $\mc D^\ast$ decides that the input $d$-separation query is false, or a state of equilibrium will eventually be reached. By definition, equilibrium is a global state of a network $G$ according to which none of the nodes in $G$ alters its state (i.e., its color) once that state is reached. This concludes the proof of Statement (II).

\subsection*{C-I.III \hspace*{10pt}Proof of Statement (III)}
\label{sec_s_I_III}
In the analysis to follow, we adhere to Assumptions (ASM-1) and (ASM-2) presented in the first paragraph of Appendix C. We analyze all potential post-equilibrium, in-transit messages.\footnote{Cast into Lamport's \emph{space-time diagram} \citep{lamport1978time}, these are the messages that cross a vertical \emph{time-cut} positioned at a (global) time which is after the occurrence of the state of equilibrium; see \citep{mattern1987algorithms}.} An in-transit message can be of {three} colors: (a1) green, (b1) red, or (c1) white. We consider each possibility next. Case (a1): If the state of equilibrium has indeed been reached, a green-colored in-transit message must be destined to a green-colored node. Indeed, if the green-colored in-transit message were destined to a red-, white-, or $\var$-colored node, it would lead, respectively, to a clash, a change in the color of the destination node, and once again, a change in the color of the destination node---all of which are in contradiction with the assumption that the state of equilibrium has already been reached. According to $\mc D^\ast$, therefore, a green-colored in-transit message will be absorbed by the corresponding destination node (which is of the same color) in time at most $\beta$ leading to the generation of no new messages. The same line of reasoning can be adopted to conclude the following (Case (b1)): A red-colored in-transit message will be absorbed by the corresponding destination node (which is of the same color) in time at most $\beta$ leading to the generation of no new messages. Next, we consider the possibility of an in-transit message being white. A white-colored in-transit message could be destined to: (a2) a white-colored node, (b2) a green-colored node, or (c2) a red-colored node. (A white-colored in-transit cannot be destined to an $\var$-colored node, as it would lead to a change in the color of the destination node---contradicting with the equilibrium assumption.) We consider each possibility in order. Case (a2): A white-colored in-transit message which is destined to a white-colored node reaches its destination in time at most $\beta$ and, according to $\mc D^\ast$, will be absorbed upon reception leading to the generation of no new messages. Case (b2): A white-colored in-transit message from node $\bb x$ to a green-colored node $\bb g$ reaches its destination, $\bb g$, in time at most $\beta$ and, according to Step (i) of $\mc D^\ast$, $\bb g$ replies, in time at most $\alpha$, by sending a green-colored message to $\bb x$ which, according to  Case (a1), will be absorbed by $\bb x$ without generating any further new messages. (Note that, according to the CUG, the receipt of a white-colored message by a green-colored node does not lead to any color update, and hence $\bb g$ does not generate any messages due to Step (ii) of $\mc D^\ast$.) Case (c2) can be handled in the same manner as Case (b2). \hfill $\blacksquare$ 

\section*{C-II\hspace*{10pt}Time-Complexity Analysis of $\mc D^\ast$}
\label{sec_appx_run_time}
We present the results in the form of two lemmas as follows. 

\textbf{Lemma C.1.} \emph{For a given DAG $G$ and disjoint sets $\bb A,\bb B,$ and $\bb C$, if $(\bb A \not\perp \bb B| \bb C)_G$ (hence, a \textsc{no}-instance $d$-separation query), then $\mc D^\ast$'s execution grants that a clash of the kind stated in Remark 1 occurs in $O(l_{An(\bb A\cup\bb B\cup\bb C)})$ time where $l_{An(\bb A\cup\bb B\cup\bb C)}$ denotes the length of the longest undirected path in the ancestral graph $G_{An(\bb A\cup\bb B\cup\bb C)}$.}

\textbf{Proof.} The proof relies on the high-level understanding of $\mc D^\ast$'s machinery as discussed in Sec.~\ref{sec_high_level}, and Statements (1) and (2) of Proposition~1 (see Sec.~C-VI of Appendix C for the proof). To obtain an upper bound on the time it takes for the clash to happen, we perform the propagation of colors through the DAG $G$ in two phases as follows. Phase-I: Starting at the nodes in $\bb C$, color white ($\wb$) propagates backwards through the DAG $G$. Phase-I ensures that all the nodes in $G$ which could potentially become white in the absence of colors red and green in the graph, indeed become white. Adopting (ASM-1) and (ASM-2) and the notation introduced therein, Phase-I is completed by time $(\alpha+\beta)l_{An(\bb A\cup\bb B\cup\bb C)}^d$ where $l^d_{An(\bb A\cup\bb B\cup\bb C)}$ denotes the longest directed path in $G_{An(\bb A\cup\bb B\cup\bb C)}$. Hence, Phase-I takes $O(l_{An(\bb A\cup\bb B\cup\bb C)}^d)$ time. Phase-II: colors green ($\gb$) and red ($\rb$) (corresponding to the nodes in $\bb A$ and $\bb B$, respectively) will be introduced back into $G$ and begin to propagate through $G$ as dictated by the machinery of $\mc D^\ast$ until along some path between a node in $\bb A$ and a node in $\bb B$ a clash takes place.\footnote{Introducing colors green ($\gb$) and red ($\rb$) back into the DAG $G$ should be interpreted as follows: Through exerting external signals, the colors of the nodes in $\bb A$ and $\bb B$ are altered to green and red, respectively. The provided interpretation is equivalent to endowing each node $\bb n\in\bb A\cup\bb B$ with a dummy child $\bb n_{ext}$ and having $\bb n_{ext}$ colored (instead of $\bb n$) green (if $\bb n\in \bb A$) or red (if $\bb n\in \bb B$) in the initialization phase of $\mc D^\ast$, thereby making any node in $\bb n\in\bb A\cup\bb B$ inherit its red/green color from its newly introduced dummy child instead of being initialized by the corresponding color in the initialization phase of $\mc D^\ast$. By this construction, we purposefully  delay the occurrence of a clash.} Adopting (ASM-1) and (ASM-2) and the notation introduced therein, after the completion of Phase-I, within time $(\alpha+\beta)l_{An(\bb A\cup\bb B\cup\bb C)}$ a clash takes place on a path between a node in $\bb A$ and a node in $\bb B$ where $l_{An(\bb A\cup\bb B\cup\bb C)}$ denote the length of the longest undirected path in $G_{An(\bb A\cup\bb B\cup\bb C)}$. Hence, putting Phase-I and Phase-II together, by time $(\alpha+\beta)(l_{An(\bb A\cup\bb B\cup\bb C)}^d+l_{An(\bb A\cup\bb B\cup\bb C)})$ a clash takes place. Note that the parameters $\alpha$ and $\beta$ are arbitrary but finite constants. Since for any DAG $G$, $l_{An(\bb A\cup\bb B\cup\bb C)}\geq l_{An(\bb A\cup\bb B\cup\bb C)}^d$, the claimed upper bound $O(l_{An(\bb A\cup\bb B\cup\bb C)})$ follows. \hfill $\blacksquare$

Using the above line of reasoning, we can prove the following lemma.

\textbf{Lemma C.2.} \emph{For a given DAG $G$ and disjoint sets $\bb A,\bb B,$ and $\bb C$, if $(\bb A\indd\bb B|\bb C)_G$ (hence, a \textsc{yes}-instance query), then $\mc D^\ast$'s execution grants that a state of equilibrium will be reached in $O(l_{An(\bb A\cup\bb B\cup\bb C)})$ time where $l_{An(\bb A\cup\bb B\cup\bb C)}$ denotes the length of the longest undirected path in ancestral graph $G_{An(\bb A\cup\bb B\cup\bb C)}$.}

{Note that, in the context of Lemma C.2, the inevitability of equilibrium state follows from Statement (1) of \tb{Proposition~1} and Statement II given in Sec.~C-I of Appendix C.}

\section*{C-III \hspace*{10pt}Proof of Proposition~2}
\label{sec_s_II_prop_2}
The analysis presented next follows the same line of reasoning presented in the proof of Lemma~C.1 in Sec.~C-II of Appendix C. For the time-complexity analysis presented blow we adhere to Assumptions (ASM-1) and (ASM-2) outlined in the first paragraph of Appendix C.\footnote{\label{footnote_cong} Note that, as the number of messages exchanged by $\mc D^\ast$ on an edge is $O(1)$ (see Statement (5) of Proposition 1), the effect of pileups (aka congestion) on a channel has been considered in Assumptions (ASM-1) and (ASM-2).} Note that both the parameters $\alpha$ and $\beta$ are arbitrary but finite constants. The claimed upper-bound $\displaystyle O\big(l_{An(\bb A\cup\bb B\cup\bb C)}^d+\min_{i,j} l_{An(\bb A\cup\bb B\cup\bb C)}^{ij}\big)$ follows from the following two statements: (s1) By time $(\alpha+\beta)l_{An(\bb A\cup\bb B\cup\bb C)}^d$, all the nodes in $G$ which could potentially become white in the absence of colors red and green in the graph, become white, and (s2) After the completion of (s1), a clash takes place on the shortest {unblocked} path between $\bb a_i$ and $\bb b_j$ in the ancestral graph $G_{An(\bb A\cup\bb B\cup\bb C)}$, within time $(\alpha+\beta)l_{An(\bb A\cup\bb B\cup\bb C)}^{ij}$, $\forall i,j$ (note that, according to the proof of Statement (1) of Proposition 1, on any unblocked path between $\bb a_i$ and $\bb b_j$ a clash eventually takes place). Hence, by time $(\alpha+\beta)(l_{An(\bb A\cup\bb B\cup\bb C)}^d+l_{An(\bb A\cup\bb B\cup\bb C)}^{ij})$ a clash will have occurred. Note that (s2) holds for all $i,j:$ $1\leq i\leq |\bb A|, 1\leq j\leq |\bb B|$. Also note that (s1) and (s2) correspond, respectively, to Phase-I and Phase-II presented in the proof of Lemma C.1 in Sec.~C-II of Appendix C. From (s1) and (s2) follows the claimed upper-bound $\displaystyle O\big(l_{An(\bb A\cup\bb B\cup\bb C)}^d+\min_{i,j} l_{An(\bb A\cup\bb B\cup\bb C)}^{ij}\big)$ in Proposition~2. \hfill $\blacksquare$ 

\section*{C-IV \hspace*{10pt}Proof of Proposition~3}
The analysis presented next follows the same line of reasoning presented in the proof of Proposition~2 (see {Sec.~C-III} for the proof). For the time-complexity analysis presented blow we adhere to Assumptions (ASM-1) and (ASM-2) outlined in the first paragraph of Appendix C; see also footnote \ref{footnote_cong}. Let $\mc M_{(\bb X \not\indd \bb Y|\bb Z)_G}$ denote a refutation-module for $(\bb X \indd \bb Y|\bb Z)_G$, with $E_{\mc M_{(\bb X \not\indd \bb Y|\bb Z)_G}}$ denoting the set of the edges of $\mc M_{(\bb X \not\indd \bb Y|\bb Z)_G}$. Let $l_{\mc M}^d$ and $|P_{\mc M}|$ denote, respectively, the length of the longest directed path and the shortest unblocked path in $\mc M_{(\bb X \not\indd \bb Y|\bb Z)_G}$. (Recall that, according to Lemma~1 in the main text, DAG $G$ must contain at least one refutation-module for $(\bb X \indd \bb Y|\bb Z)_G$, and, due to Definition~1 in the main text, $\mc M_{(\bb X \not\indd \bb Y|\bb Z)_G}$ must contain at least one unblocked path between a node in $\bb X$ and a node in $\bb Y$.) Also, let $\mc M_{(\bb X \not\indd \bb Y|\bb Z)_G}^{\ast}$ denote the minimal refutation-module for $(\bb X \indd \bb Y|\bb Z)_G$, with $E_{\mc M_{(\bb X \not\indd \bb Y|\bb Z)_G}^{\ast}}$ denoting the set of the edges of $\mc M_{(\bb X \not\indd \bb Y|\bb Z)_G}^{\ast}$, and $l_{\mc M^\ast}^d$ and $|P_{\mc M^\ast}|$ denoting the length of the longest directed path and the shortest unblocked path in $\mc M_{(\bb X \not\indd \bb Y|\bb Z)_G}^\ast$, respectively. The claimed upper-bound $O(\min_{\mc M_{(\bb X \not\indd \bb Y|\bb Z)_G}} \{l_{\mc M}^d+|P_{\mc M}|\})$ follows from the following two statements: (s1) By time $(\alpha+\beta)l_{\mc M}^d$, all the nodes in $\mc M_{(\bb X \not\indd \bb Y|\bb Z)_G}$ which could potentially become white in the absence of colors red and green in the graph, become white, and (s2) After the completion of (s1), a clash takes place along the shortest unblocked path $P_{\mc M}$ in $\mc M_{(\bb X \not\indd \bb Y|\bb Z)_G}$ within $(\alpha+\beta)|P_{\mc M}|$. (Note that (s1) and (s2) correspond, respectively, to Phase-I and Phase-II presented in the proof of Lemma C.1 in Sec.~C-II of Appendix C.) Hence, taken together, a clash will have occurred along $P_{\mc M}$ by time $(\alpha+\beta)(l_{\mc M}^d+|P_{\mc M}|)$. Since the above argument holds for any arbitrary refutation-module $\mc M_{(\bb X \not\indd \bb Y|\bb Z)_G}$, it follows that a clash will have occurred by time $(\alpha+\beta)\min_{\mc M_{(\bb X \not\indd \bb Y|\bb Z)_G}}\{l_{\mc M}^d+|P_{\mc M}|\}$, hence the claimed upper-bound $O(\min_{\mc M_{(\bb X \not\indd \bb Y|\bb Z)_G}} \{l_{\mc M}^d+|P_{\mc M}|\})$ in Proposition~3. Finally, since $\min_{\mc M_{(\bb X \not\indd \bb Y|\bb Z)_G}}(l_{\mc M}^d+|P_{\mc M}|)\leq l_{\mc M^\ast}^d+|P_{\mc M^\ast}|$, $l_{\mc M^\ast}^d\leq |E_{\mc M_{(\bb X \not\indd \bb Y|\bb Z)_G}^\ast}|$, and $|P_{\mc M^\ast}|\leq |E_{\mc M_{(\bb X \not\indd \bb Y|\bb Z)_G}^\ast}|$, it follows that $\min_{\mc M_{(\bb X \not\indd \bb Y|\bb Z)_G}} \{l_{\mc M}^d+|P_{\mc M}|\}\leq O(|E_{\mc M_{(\bb X \not\indd \bb Y|\bb Z)_G}^{\ast}}|)$, hence the claimed upper-bound $O(|E_{\mc M_{(\bb X \not\indd \bb Y|\bb Z)_G}^{\ast}}|)$ on the time for the occurrence of a clash. This concludes the proof. \hfill $\blacksquare$

\section*{C-V \hspace*{10pt}Proof of Proposition~4}
Consider any refutation-module $\mc M_{(\bb X \not\indd \bb Y|\bb Z)_G}^{\dagger}$ satisfying the following condition: $\mc M_{(\bb X \not\indd \bb Y|\bb Z)_G}^{\dagger}$ contains the unblocked path $\min_{i,j} l_{An(\bb X\cup\bb Y\cup\bb Z)}^{ij}$. By definition, it immediately follows that $|P_{\mc M^\dagger}|\leq \min_{i,j} l_{An(\bb X\cup\bb Y\cup\bb Z)}^{ij}$ and $l_{\mc M^\dagger}^d \leq l_{An(\bb X\cup\bb Y\cup\bb Z)}^d$; for the notation, see Propositions~2 and 3 in the main text. Hence, $l_{\mc M^\dagger}^d+|P_{\mc M^\dagger}|\leq l_{An(\bb X\cup\bb Y\cup\bb Z)}^d + \min_{i,j} l_{An(\bb X\cup\bb Y\cup\bb Z)}^{ij}$. Since $\min_{\mc M_{(\bb X \not\indd \bb Y|\bb Z)_G}}\{l_{\mc M}^d+|P_{\mc M}|\}\leq l_{\mc M^\dagger}^d+|P_{\mc M^\dagger}|$, the claim of Proposition~4 follows. \hfill $\blacksquare$

\section*{C-VI\hspace*{10pt} Proof of Proposition~1}
\label{apx_prop_1_proof}

\subsubsection*{Proof of Statement~(1)}  
The reader is referred to {Sec.~C-I.I} of Appendix C for the proof.

\subsubsection*{Proof of Statement~(2)}
Let $\bb I_T$ denote the set of all nodes $\bb i$ in $G$ which have ever sent a message up to a (global) time  $T$. Then, the validity of Statement (2) follows from the following recursion. If \textbf{($\dagger$)} Prior to applying Steps (i) and (ii) on the corresponding recipients of $\bb i$'s messages, $\bb i\in \bb I_T$, the set $\bb I_T$ satisfies the following two conditions: ($\star$) Any node in the set belongs to $G_{An(\bb A\cup\bb B\cup\bb C)}$, and ($\star\star$) Any node to which a member of the set has ever sent a message belongs to $G_{An(\bb A\cup\bb B\cup\bb C)}$, then \textbf{($\ddagger$)} After applying Steps (i) and (ii) on the corresponding recipients of $\bb i$'s messages (denoted by the set $recp(\bb i)$) for all $\bb i\in \bb I_T$, the set $(\cup_{\bb i\in \bb I_T}recp(\bb i))\cup \bb I_T$ indeed satisfies ($\star$) and ($\star\star$). Note that at the initial configuration of $\mc D^\ast$, nodes in $\bb A\cup\bb B\cup\bb C$ send their corresponding colors to their parents. Hence, Statement ($\dagger$) holds at the initial configuration. Let us now assume that Statement ($\dagger$) holds for the set of all nodes $\bb i$ in $G$ that have ever sent a message up to a (global) time $T$, i.e., $\bb i\in \bb I_T$. According to Step (i) of $\mc D^\ast$, a node $\bb x$ which is the recipient of a message from $\bb i$ replies back by sending its own color to the sender. Note that, according to Statement ($\dagger$), the sender must adhere to ($\star$) and ($\star\star$).  According to Step (ii) of $\mc D^\ast$, the node $\bb x$ sends its updated color to (a) all its parents, and (b) those children of $\bb x$ with which $\bb x$ has communicated before. Based on the argument provided above regarding Step (i), $\bb x$ sending messages to its parents guarantees that the set $(\cup_{\bb i\in \bb I_T}recp(\bb i))\cup\bb I_T$ indeed satisfies ($\star$) and ($\star\star$). Also, due to the constraints ``with which $\bb x$ has communicated before" in (b) and Statement ($\star\star$), $\bb x$ sending messages to the nodes prescribed in (b) guarantees that the set $(\cup_{\bb i\in \bb I_T}recp(\bb i))\cup\bb I_T$ indeed satisfies ($\star$) and ($\star\star$). The above argument establishes the validity of the recursion. The recursion given above, together with the fact that Statement ($\dagger$) holds at the initial configuration of $\mc D^\ast$, grants the validity of Statement~(2). This concludes the proof. \hfill $\blacksquare$

\subsubsection*{Proof of Statement~(3)}
\tb{Statement (3) follows from Statement (I) (see {Sec.~C-I.I} of Appendix C), Lemma C.1 and Lemma C.2 (see Sec.~{C-II} of Appendix C).} 

\subsubsection*{Proof of Statement~(4)}
\tb{The reader is referred to {Sec.~C-I.III} of Appendix C for the proof.}

\subsubsection*{Proof of Statement~(5)}
In what follows we prove that the number of messages exchanged on an edge is bounded above by a constant which is independent of the size of the graph. Based on Step (ii) of $\mc D^\ast$, a node sends out a message upon updating its color and discovering that its new color is different from its {pre-update} color. According to the state transition diagram depicted in Fig.~\ref{fig_transition}, throughout an execution of $\mc D^\ast$ a node could pass through at most five states $\{\var,\wb,\gb,\rb,clash\}$, and once a node changes its state it cannot go back to that state ever again (due to the nonexistence of any loop of the length at least two in the state transition diagram, see Fig.~\ref{fig_transition}). Hence, Step (ii) of $\mc D^\ast$ results in $O(1)$ messages to be exchanged per channel. There are ${5\choose 2} = 10$ possible ways of pairing nodes of different colors together. By inspection, Step (i) of $\mc D^\ast$ results in having the highest number of messages exchanged on the edge between a white-colored node and a green-colored node, which, as we show, results in $O(1)$ messages to be exchanged on that edge. When a (newly) white-colored node $\bb p$ sends a white-colored message to its green-colored neighbor $\bb q$ the following exchange of messages takes place due to Step (i): white-colored message (sent by the white-colored initiator node $\bb p$) will be received by the green-colored node $\bb q$; $\bb q$ will send back a green-colored message (due to Step (i)). Upon receipt of the green-colored message by $\bb p$, $\bb p$ will send a new white-colored message to $\bb q$ (due to Step (i)) and also will updates its color to green based on the CUG. Upon receipt of the new white-colored message, $\bb q$ will send back a green-colored message (due to Step (i)) to $\bb p$. However, since $\bb p$'s color has been updated to green (hence,  identical to that of the received message), this time $\bb p$ does not send any message to $\bb q$ due to Step (i). This completes the proof. \hfill $\blacksquare$ 

\subsubsection*{Proof of Statement~(6)}
Throughout an execution of $\mc D^\ast$ there exist three types of messages which can be exchanged between nodes, namely, a white-, a green-, or a red-colored message. Therefore, to encode the said three types two bits are required. Statement~(6) then follows from Statement (5) and the argument provided above. \hfill $\blacksquare$ 

\section*{C-VII\hspace*{10pt} Alternative, Centralized Initialization of $\mc D^\ast$}
Below, we explain how the initialization phase of $\mc D^\ast$ can be accomplished, in a distributed manner, in $O(l)$ time, where $l$ denotes the length of the longest undirected path in $G$. 
 
First, using a combination of broadcast and convergecast, a pre-assigned initiator node, $\bb s$, sends the initialization message \tb{$\sdbc{\texttt{INITIALIZE}}$} (containing the list of nodes in sets $\bb A$, $\bb B$, and $\bb C$) to all the other nodes in $G$, and receives an acknowledgement that all nodes have received \tb{$\sdbc{\texttt{INITIALIZE}}$}. Using the \emph{AsynchSpanningTree} algorithm in \citep{lynch1996distributed}, this can be done in time $O(l)$; cf. \citep[p. 499, 2nd paragraph]{lynch1996distributed}.
 
Then, $\bb s$ broadcasts the control message, \tb{$\sdbc{\texttt{START}_{\mc D^\ast}}$}, to all the nodes in $G$. Using the \emph{AsynchSpanningTree} algorithm, this can be done in time $O(D)$, where $D$ denotes the diameter of $G$. Upon receipt of \tb{$\sdbc{\texttt{START}_{\mc D^\ast}}$} by a node in $\bb A\cup\bb B\cup\bb C$, it sends its color to its parents, as prescribed in the initialization phase of $\mc D^\ast$ outlined in Sec.~\ref{sec_alg_main} in the main text.

\section*{C-VIII\hspace*{10pt} On the Order-Invariance Property}
\label{sec_order_inv}
Before we state the result in a form of a lemma, let us introduce the following notation. We adopt the expression $S_i^{\bb x}\overset{M_1,\cdots,M_n}{\leadsto}S_f^{\bb x}$ to state the following: Starting at the state $S_i^{\bb x}$, node $\bb x$ transits to the state $S_f^{\bb x}$ upon receiving the sequence of messages $M_1,\cdots,M_n$ where $M_1,\cdots,M_n \in \{\wb,\gb,\rb\}$ and $S_i^{\bb x}, S_f^{\bb x}\in\{\var,\wb,\gb,\rb,clash\}$ . Let us now formally state the result as a lemma.

\textbf{Lemma C.3.}
\emph{Let} $\bb x$ \emph{be a node in the network. Then, the following holds:} 
\begin{eqnarray*}
\displaystyle (S_i^{\bb x}\overset{M_1,M_2}{\leadsto}S_f^{{\bb x}})\Rightarrow  (S_i^{\bb x}\overset{M_2,M_1}{\leadsto}S_f^{\bb x}).
\end{eqnarray*}

\textbf{Proof.} The proof can be straightforwardly accomplished by examining all the possible cases and showing that the statement holds true for all of them. To provide a sample case, consider $\wb\overset{\wb,\gb}{\leadsto}\gb$. Using the CUG given in Sec.~\ref{sec_alg_main}, it is straightforward to check that $\wb\overset{\gb,\wb}{\leadsto}\gb$ holds true. \hfill $\blacksquare$

The order-invariance property is captured in the following lemma. 

\textbf{Lemma C.4}
\emph{Let} $\bb x$ \emph{be a node in the network and let $\pi$ be an arbitrary permutation defined on the set $\{1,2,\cdots,n\}$. Then, the following holds:} 
\begin{eqnarray*}
(S_i^{\bb x}\overset{M_1,\cdots,M_n}{\leadsto}S_f^{\bb x})\Rightarrow (S_i^{\bb x}\overset{M_{\pi(1)},\cdots,M_{\pi(n)}}{\leadsto}S_f^{\bb x})
\end{eqnarray*}

\textbf{Proof.} The proof follows from Lemma C.3 and the understanding that the sequence\linebreak $M_{\pi(1)},\cdots,M_{\pi(n)}$ (for an arbitrary permutation $\pi$) can be constructed from the original sequence $M_1,\cdots,M_n$ through a series of pairwise permutations.\footnote{This is essentially the idea behind the well-known sorting algorithm, Insertion Sort.} \hfill $\blacksquare$

It is worth noting that the order-invariance property formalized above is analogous to the key notion of \emph{exchangeability} in probability theory.

\nocite{griffiths2015rational}
\nocite{Nobandegani2017phdthesis}
\nocite{icard2015}
\nocite{Nobandegani2017frame}
\nocite{ma2014neural}
\nocite{ma2008spiking}
\nocite{summerfield2015humans}

\renewcommand\bibliographytypesize{\normalsize}
\bibliographystyle{apacite}
\bibliography{ref_d_sep_arxiv}
\end{document}